\documentclass[a4paper,fleqn]{cas-dc}

\usepackage[numbers,sort&compress,longnamesfirst]{natbib}
\usepackage{epstopdf}
\usepackage{color}
\usepackage{bm}
\usepackage{amsmath}
\usepackage{stfloats}  
\usepackage{subfigure}
\usepackage{float}
\usepackage{caption}
\usepackage{multirow}
\usepackage{algorithm}
\usepackage{algorithmicx}
\usepackage{algpseudocode}
\usepackage{pifont}
\usepackage[normalem]{ulem}
\useunder{\uline}{\ul}{}
\makeatletter
\renewcommand{\@thesubfigure}{\hskip\subfiglabelskip}
\makeatother

\def\tsc#1{\csdef{#1}{\textsc{\lowercase{#1}}\xspace}}
\tsc{WGM}
\tsc{QE}
\tsc{EP}
\tsc{PMS}
\tsc{BEC}
\tsc{DE}

\begin{document}
	\let\WriteBookmarks\relax
	\def\floatpagepagefraction{1}
	\def\textpagefraction{.001}
	\shorttitle{}
	\shortauthors{X. Shang et~al.} 
	\title [mode = title]{Holistic Dynamic Frequency Transformer for Image Fusion and Exposure Correction}

	\author[1]{\textcolor[RGB]{0,0,1}{Xiaoke Shang}}
	\author[2]{\textcolor[RGB]{0,0,1}{Gehui Li}}
	\author[3,4]{\textcolor[RGB]{0,0,1}{Zhiying Jiang}}
	\author[1]{\textcolor[RGB]{0,0,1}{Shaomin Zhang}}
	\author[1]{\textcolor[RGB]{0,0,1}{Nai Ding}}
	\author[5]{\textcolor[RGB]{0,0,1}{Jinyuan Liu}}[orcid=0000-0003-2085-2676]
	\cormark[1]%
	\address[1]{School of Biomedical Engineering and Instrument Science, Zhejiang University, Hangzhou 310027, China}
	\address[2]{DUT-RU International School of Information Science $\&$ Engineering, Dalian University of Technology, Dalian 116620, China}
	\address[3]{School of Software Technology, Dalian University of Technology, Dalian 116024, China}
	\address[4]{Key Laboratory for Ubiquitous Network and Service Software of Liaoning Province, Dalian 116620, China}
	\address[5]{School of Mechanical Engineering, Dalian University of Technology, Dalian 116086, China}
	\ead{atlantis918@hotmail.com}
	\cortext[cor1]{Corresponding author: } 

	\begin{abstract}
The correction of exposure-related issues is a pivotal component in enhancing the quality of images, offering substantial implications for various computer vision tasks. Historically, most methodologies have predominantly utilized spatial domain recovery, offering limited consideration to the potentialities of the frequency domain. Additionally, there has been a lack of a unified perspective towards low-light enhancement, exposure correction, and multi-exposure fusion, complicating and impeding the optimization of image processing. In response to these challenges, this paper proposes a novel methodology that leverages the frequency domain to improve and unify the handling of exposure correction tasks. Our method introduces Holistic Frequency Attention and Dynamic Frequency Feed-Forward Network, which replace conventional correlation computation in the spatial-domain. They form a foundational building block that facilitates a U-shaped Holistic Dynamic Frequency Transformer as a filter to extract global information and dynamically select important frequency bands for image restoration. Complementing this, we employ a Laplacian pyramid to decompose images into distinct frequency bands, followed by multiple restorers, each tuned to recover specific frequency-band information. The pyramid fusion allows a more detailed and nuanced image restoration process. Ultimately, our structure unifies the three tasks of low-light enhancement, exposure correction, and multi-exposure fusion, enabling comprehensive treatment of all classical exposure errors. Benchmarking on mainstream datasets for these tasks, our proposed method achieves state-of-the-art results, paving the way for more sophisticated and unified solutions in exposure correction.
	\end{abstract}
	
	


	\maketitle  

\section{Introduction}
Exposure-related tasks, including low-light enhancement, exposure correction, and multi-exposure fusion, bear significant implications in the field of computer vision. Proper image exposure is instrumental in achieving high-quality visual information, making it pivotal for effective image analysis and processing. Overexposure and underexposure, both common imaging issues, lead to loss of details and reduced contrast, thereby hindering the visual appeal and practical usability of images.

The evolution of methodologies to address low-light enhancement has seen significant advancements over the years. Initial methods relied on traditional image processing techniques, including histogram equalization and gamma correction. These were soon replaced by sophisticated techniques leveraging deep learning. Notable among these are Deep-UPE \cite{wang2019underexposed} and Zero-DCE \cite{guo2020zero}, which optimized for local contrast enhancement in underexposed images. However, these models primarily focused on underexposure correction, leaving overexposed images relatively unaddressed.
This led to the expansion of low-light enhancement techniques into exposure correction tasks, an evolution marked by the introduction of the Multi-Scale Exposure Correction (MSEC) \cite{afifi2021learning} method, which targets both over- and underexposed regions in images. LCDP \cite{wang2022local} is proposed to rectify multiple exposure errors present within the same image. FECNet \cite{huang2022deep}, a Fourier and convolution-based Exposure Correction Network, comprises an amplitude sub-network and a phase sub-network, which restore the amplitude and phase representations respectively. Existing exposure correction networks have not adequately addressed the issues of color and detail loss during the correction process. Moreover, a comprehensive integration and application of Transformer networks with frequency domain approaches have not been explored. The task generalization of existing schemes also remains a pressing issue that needs to be resolved.

As for the multi-exposure fusion tasks, the prevailing solutions can be generally divided into several categories. Traditional methods \cite{GFF, liu2015dense} like gradient-based fusion and pyramid-based fusion provide a foundation but often result in artifacts or insufficient detail preservation. More recent contributions have capitalized on the power of deep learning, such as the fusion algorithm based on a convolutional neural network \cite{Deepfuse, PMGI, DPE-MEF, zhang2023multi, tang2023rethinking, hu2023zmff, tang2023divfusion}. While they advanced the performance of fusion tasks significantly, they still lacked a holistic approach to addressing exposure-related tasks in their entirety.

Our proposed method intervenes to overcome existing limitations by introducing an innovative, unified framework for exposure correction that spans the domains of low-light enhancement, exposure correction, and multi-exposure fusion. 
Recognizing that low-light data serves as a subset of exposure-error data, which includes both overexposure and underexposure, and multi-exposure fusion often necessitates subsequent detail and color restoration and enhancement, our approach centers on devising a robust, universal exposure corrector engineered to harmonize these three intertwined tasks into a cohesive solution.

We propose Holistic Frequency Attention (HFA), and Dynamic Frequency Feed-Forward Network (DFFFN). They form a U-shaped restorer that establishes global dependencies and dynamically filters the main information according to the frequency band. Paired with our Laplacian pyramid for image decomposition and pyramid fusion, the approach offers fine-grained and nuanced image restoration, resulting in enhanced detail and color correction.

The contributions of this paper are as follows:
\begin{itemize}
\item We introduce a U-shaped Restorer that combines Holistic Frequency Attention and Dynamic Frequency Feed-Forward Network, designed to efficiently extract global features and dynamically filter crucial frequency bands, thereby achieving enhanced detail and color in exposure-corrected images.
\item Our method leverages Laplacian pyramids to decompose images and employs multiple restorers for pyramid fusion, enhancing the information synthesis from different frequency bands.
\item We unveil an integrated architecture that synergistically addresses low-light enhancement, exposure correction, and multi-exposure fusion tasks by leveraging the inherent correlations among these three challenges, demonstrating its versatility in tackling a wide range of exposure-related issues.
\item Our method achieves state-of-the-art performance on both downstream and upstream tasks, demonstrating its effectiveness and applicability.
\end{itemize}

\section{Related Work}
\subsection{Low-Light Enhancement and Exposure-Error Correction}
Low-light image enhancement has been an active research topic in computer vision due to its importance in improving the visual quality of images and the performance of visual recognition tasks under poor lighting conditions. Methods for low-light image enhancement can be broadly categorized into three types: histogram-based methods, Retinex-based methods, and deep learning-based methods.

Initially, research predominantly focused on histogram-based methods, which offered a computationally efficient and straightforward approach. These methods typically manipulate the histogram of input images, stretching the contrast to boost the visibility of features under low-light conditions. Seminal works like the Dynamic Histogram Equalization (DHE) by Abdullah and Fofi \cite{abdullah2007dynamic}, and the Adaptive Histogram Equalization (AHE) by Pizer et al. \cite{pizer1987adaptive}, exemplify this approach. In parallel, Reza \cite{reza2004realization} proposed an efficient realization of histogram equalization, while Ying et al. \cite{ying2017new} introduced the contrast-limited adaptive histogram equalization (CLAHE) technique, specifically designed to suppress noise over-amplification.

Building upon these foundational methods, subsequent researchers began exploring the potential of the Retinex theory. This theory involves the separation of an image into illumination and reflectance components, enabling finer control over image enhancement. Influential contributions include the Retinex model by Land and McCann \cite{land1977retinex}, and the low-light image enhancement (LIME) method by Guo et al. \cite{guo2016lime}. Over time, more sophisticated Retinex-based models were developed, such as the robust Retinex model by Li et al. \cite{li2018structure}, the low-rank regularized Retinex model (LR3M) by Ren et al. \cite{ren2020lr3m}, and the joint intrinsic-extrinsic prior model by Cai et al. \cite{cai2017joint}.

With the advent of deep learning, researchers have found a potent tool for tackling low-light enhancement \cite{ma2023bilevel, liu2020real}. Chen et al. \cite{chen2018deep} effectively harnessed two-way generative adversarial networks (GANs) in an unpaired learning method for image enhancement. In a similar vein, Lv et al. \cite{lv2021attention} utilized a multi-branch convolutional neural network in an end-to-end attention-guided approach. Other research, such as that by Wang et al. \cite{wang2019underexposed}, explored neural networks' capabilities in enhancing underexposed photos by introducing intermediate illumination into the network. Wei et al. \cite{wei2018deep} proposed unique solutions by integrating signal prior-guided layer separation, data-driven mapping networks, and deep Retinex-Nets for low-light enhancement.

Recent studies in the field have expanded beyond low-light enhancement to address exposure correction \cite{ma2022practical}, tackling both over- and under-exposed images. Afifi et al. \cite{afifi2021learning} laid the groundwork by splitting the exposure correction into color and detail enhancement tasks and using a wide-ranging exposure dataset. Building on this, Cui et al. \cite{cui2022illumination} developed the Illumination Adaptive Transformer (IAT) to adjust color correction and gamma correction parameters in images under various lighting conditions. Complementing these efforts, Wang et al. \cite{wang2022local} proposed a method exploiting local color distributions to enhance regions suffering from both over- and under-exposure, introducing a dual-illumination learning mechanism and a new dataset to aid this process.

\begin{figure*}[!htb]
\centering
\setlength{\tabcolsep}{1pt}
\begin{tabular}{c}
\includegraphics[width=0.95\textwidth]{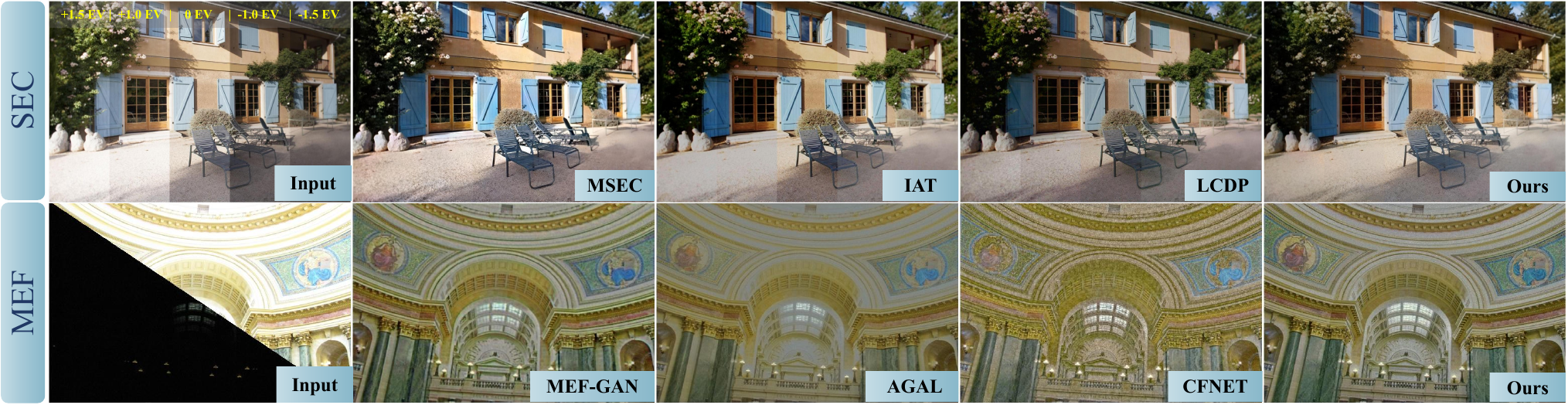}
\end{tabular}
\caption{The proposed method produce a single-exposure-corrected(SEC)/multiple-exposure-fused(MEF) image with clear details and visually pleasing colors.}
\label{fig:topfigure}
\end{figure*}

\subsection{Multi-Exposure Fusion}
The development of Multi-Exposure Image Fusion \cite{guo2022image, tang2022piafusion, han2022multi, wang2023interactively, liu2023holoco, liu2021learning, liu2020bilevel, liu2021smoa, liu2021searching, huang2022reconet, liu2022learning, ma2022swinfusion, ma2022low, liu2022coconet, liu2018learning} has seen many significant strides, primarily utilizing deep learning methodologies.
Kalantari et al. \cite{kalantari2017deep} utilized a Convolutional Neural Network (CNN) to merge HDR images, mitigating ghosting and tearing artifacts often appearing in dynamic scenes.
Contrastingly, Ma et al. \cite{ma2019deep} proposed MEF-Net, a swift MEF method employing a fully convolutional network for weight map prediction, outpacing and outperforming many traditional methods.
Yin et al. \cite{yin2020deep} integrated both pixel-level and feature-level considerations into a novel encoder-decoder network, ensuring fine-grained control, semantic consistency, and texture calibration.
Ram et al. \cite{ram2017deepfuse} tackled the limitations of hand-crafted feature-based MEF methods by introducing an unsupervised deep learning architecture, demonstrating superior performance on natural images.
Xu et al. \cite{xu2020mef} introduced MEF-GAN, an adversarial network incorporating self-attention mechanism, allowing for effective correction of local distortion and inappropriate representation.
Chen et al. \cite{chen2020deep} proposed a network that integrates multiple mechanisms such as homography estimation, attention mechanism, and adversarial learning to address ghosting issues and misalignment.
Most recently, Liu et al. \cite{liu2022attention} developed an attention-guided global-local adversarial learning network, ensuring the alignment of local patches of the fused images with realistic normal-exposure ones, thereby restoring realistic texture details and correcting color distortion.

Existing methodologies in the field of Multi-Exposure Image Fusion (MEF) often separately study exposure correction and multi-exposure fusion, with no single method unifying these three tasks. Recognizing the correlation between multi-exposure fusion and exposure correction, our proposed method brings these two aspects under a unified network. By concurrently addressing these tasks, our approach seeks to leverage the interdependencies between them, ultimately aiming to improve the overall performance and efficacy of multi-exposure image fusion.

\subsection{Fourier Transform in Computer Vision}
Fourier Transform has played a pivotal role in advancing computer vision tasks \cite{piao2019depth, piao2020a2dele, zhang2020select, liu2012fixed, wu2019essential}. By transferring an image from its spatial domain to the frequency domain, it allows for the identification and extraction of informative features that are often more resilient to local variations and noise. 
Rao et al. \cite{rao2021global} introduced the Global Filter Network (GFNet), a model that learns spatial dependencies in the frequency domain using 2D discrete Fourier transforms, demonstrating excellent accuracy and efficiency.
Xu et al. \cite{xu2021fourier} presented a Fourier-based approach for domain generalization tasks. By leveraging Fourier phase information, they developed an effective data augmentation strategy and a co-teacher regularization technique.
Chi et al. \cite{chi2020fast} proposed the Fast Fourier Convolution (FFC) operator, an innovative design encapsulating different scales of computations within a single unit.
Kong et al. \cite{kong2023efficient} proposed frequency domain attention utilizing Fast Fourier Transform to reduce attention complexity, and employed a gated discriminative filtering mechanism to address the deblurring problem.

Despite the success of Fourier Transform in numerous computer vision tasks, there is a noticeable gap in its application for exposure correction and multi-exposure fusion tasks. These tasks have yet to be deeply researched from the frequency domain perspective. 

\begin{figure}[h]
\centering
\setlength{\tabcolsep}{1pt}
\begin{tabular}{c}
\includegraphics[width=0.9\columnwidth]{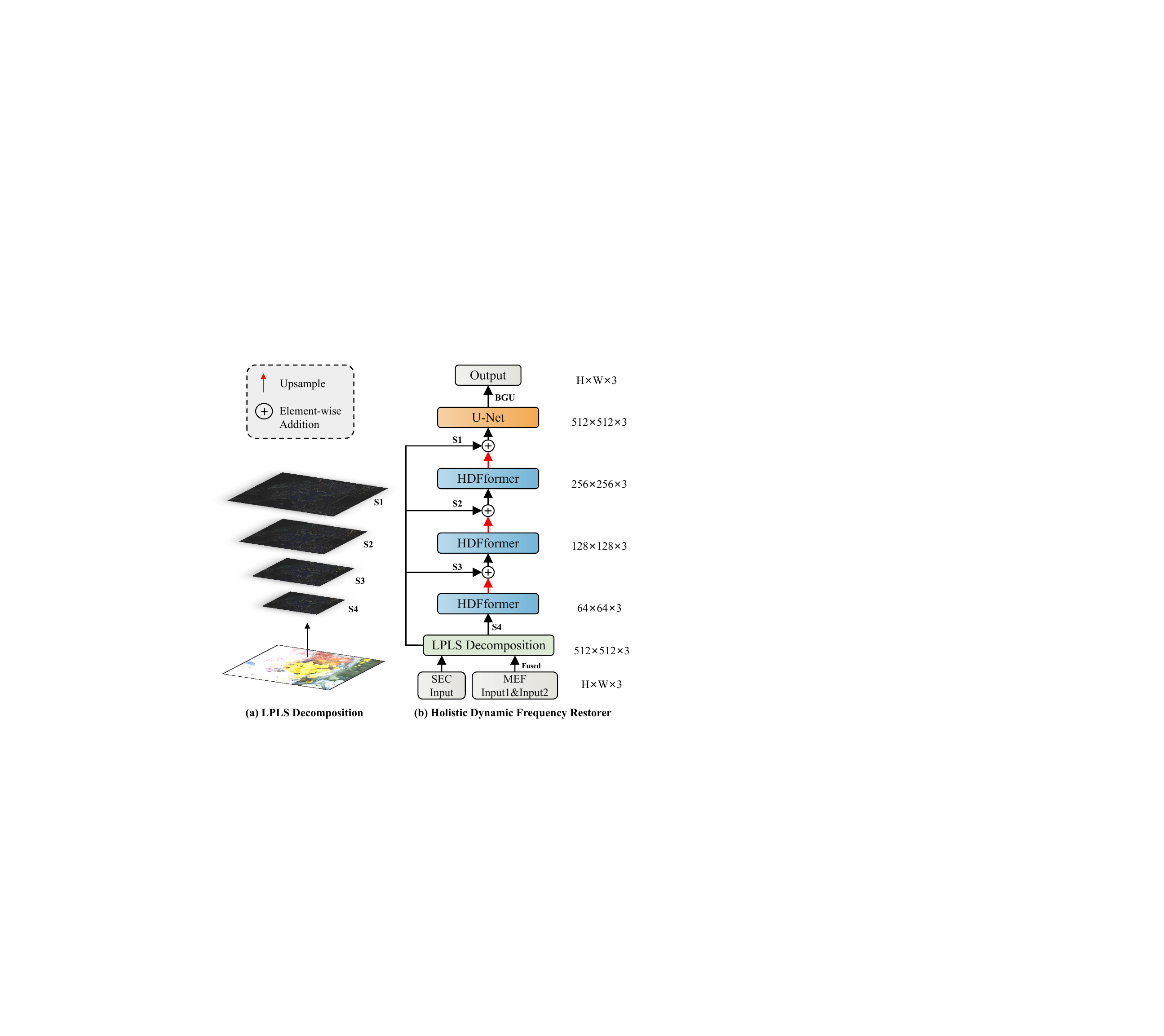}
\end{tabular}
\caption{The workflow of the proposed Restorer.}
\label{fig:workflow}
\end{figure}

\begin{figure*}[!htb]
\centering
\setlength{\tabcolsep}{1pt}
\begin{tabular}{c}
\includegraphics[width=0.86\textwidth]{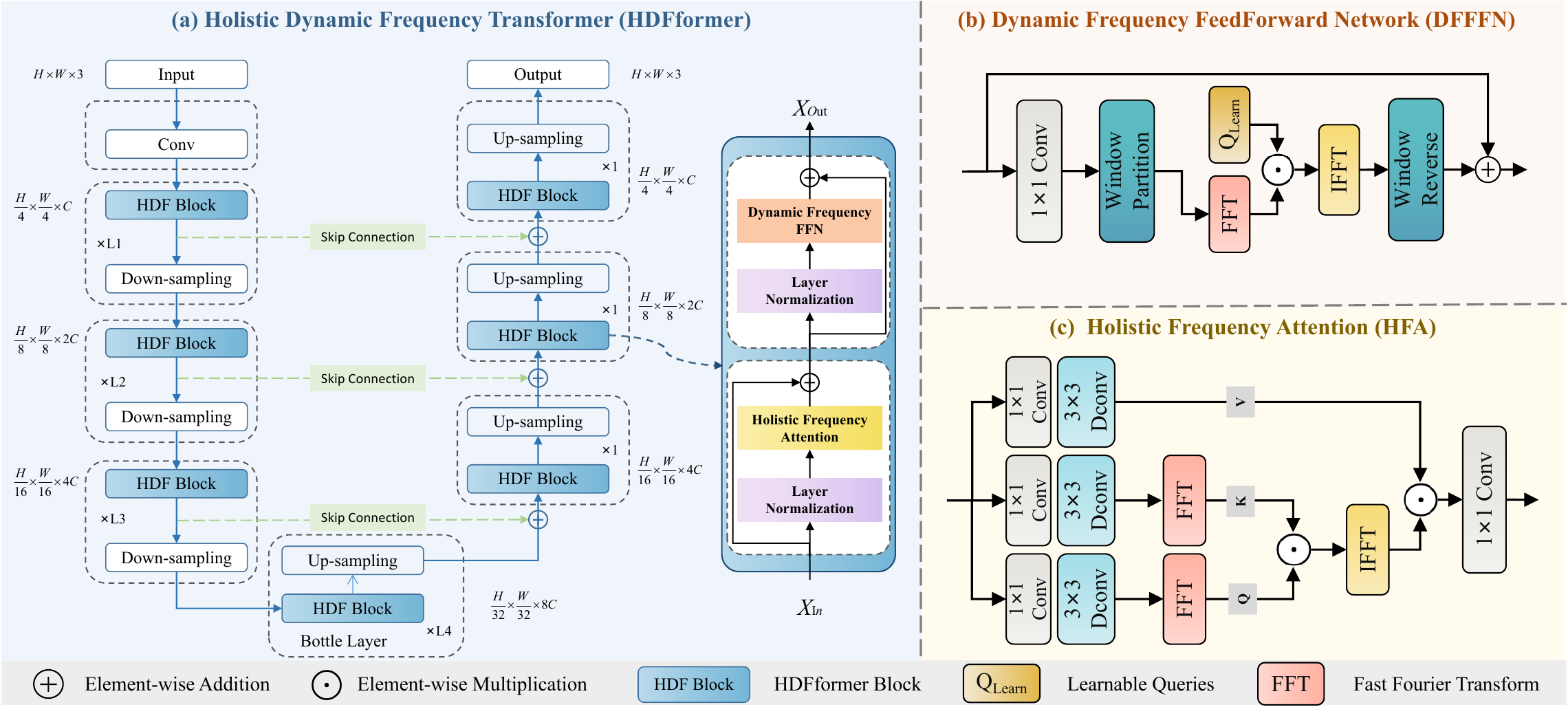}
\end{tabular}
\caption{The workflow of the proposed Holistic Dynamic Frequency Transformer.}
\label{fig:restorer}
\vspace{-20pt}
\end{figure*}

\section{Proposed Method}
We utilize Laplacian pyramids to decompose the images into different frequency bands, then employ multiple restorers, each of which restores information specific to a particular frequency band, thereby facilitating pyramid fusion. This multi-level decomposition and fusion strategy enables more detailed and comprehensive extraction and synthesis of image information from different frequency bands, enhancing the quality of the output images. Figure \ref{fig:workflow} shows the workflow of the proposed method. In the following, we present the details of each component.

\subsection{Laplacian Pyramid Decomposition}
The principle of image pyramid is that each image is decomposed into N layers of multi-scale pyramid image sequence. The image with the smallest size is taken as the 1st layer and the image with the largest size is taken as the Nth layer. The size of the image in the Kth layer is one-fourth of the size of the image in the K+1th layer. During the operation of Gaussian pyramid, the image is filtered by Gaussian blur and down-sampling operation will lose some high frequency detail information. To describe this high-frequency information, Laplace pyramid is defined. The image of each layer of the Gaussian pyramid is subtracted from the image of the next layer after up-sampling and Gaussian filtering. A series of difference images are obtained, which are the images after the Laplacian pyramid decomposition \cite{afifi2021learning}. Mathematically defined as: $\mathrm{L}_{i}=\mathrm{G}_{\mathrm{i}}-\operatorname{PyrUp}\left(\mathrm{G}_{\mathrm{i+1}}\right)$. The purpose of this operation is to decompose the source images into different spatial frequency bands, so that separate networks can be used to restore features and details of specific frequency bands at different decomposition layers.

\subsection{Holistic Dynamic Frequency Transformer}
By leveraging multiplication in the frequency domain to replace correlation calculation in the temporal domain, we introduce a novel attention mechanism and FFN layer that are based on the frequency domain. These two components form a building block, which then constitutes a U-shaped restorer, functioning as a filter to selectively restore the required frequency domain for image reconstruction. This innovative concept enhances the computational efficiency and performance of the attention mechanism in computer vision tasks. Figure \ref{fig:restorer} shows the workflow of the proposed restorer.

\subsubsection{Holistic Frequency Attention}
Our work primarily focuses on the Window Attention mechanism in the context of computer vision. Window Attention is a variant of the attention mechanism where computations are restricted within local windows, which are subparts of the input image or feature map.

In the Attention mechanism, input $X$ is multiplied by three mapping matrices $W_q$, $W_k$, and $W_v$ to generate the query $\mathbf{Q}$, key $\mathbf{K}$, and value $\mathbf{V}$ respectively:
\begin{equation}
\mathbf{Q} = XW_q, \quad \mathbf{K} = XW_k, \quad \mathbf{V} = XW_v.
\end{equation}

The formulation of Attention is as follows:
\begin{equation}
A = Softmax(\mathbf{QK^T}/\sqrt{d_k})\mathbf{V},
\end{equation}
where $d_k$ is the dimensionality of the query and key vectors, and $A$ is the output attention map. 

The time complexity of this operation is $O(N^2)$, where $N$ is the number of pixels in the image. The limitation of attention is its computational complexity. When the image size increases, the required computations increase quadratically. Window attention is used to reduce complexity. Its complexity is $O(DN)$, where D is the number of pixels in the window. On the one hand, its complexity deteriorates to quadratic as the window increases. On the other hand, the partitioning into windows can result in loss of long-range dependencies because the attention is only applied within individual windows.

An inspired insight from \cite{kong2023efficient} leverages the classical Convolution Theorem, demonstrating that multiplication in the frequency domain can effectively replace correlation operations in the spatial domain.
Accordingly, we propose to replace the matrix multiplication between attention query and key with multiplication in the frequency domain without windowing.
Furthermore, from the perspective of \cite{chi2020fast}, this can be interpreted as query and key mutually filtering each other in the frequency domain.
The specific process is as follows: First, we use the Fast Fourier Transform (FFT) to transform the query and key into the frequency domain, represented as $Q_f$ and $K_f$:
\begin{equation}
Q_f = FFT(\mathbf{Q}), \quad K_f = FFT(\mathbf{K}).
\end{equation}

We then perform element-wise multiplication in the frequency domain:
\begin{equation}
M_f = Q_f \odot K_f,
\end{equation}
where $\odot$ denotes element-wise multiplication. Subsequently, we use the Inverse Fast Fourier Transform (IFFT) to transform back to the spatial domain:
\begin{equation}
M = IFFT(M_f).
\end{equation}

The attention matrix $A$ and value $V$ are then subject to a Hadamard product, giving the final output after attention:
\begin{equation}
A = Softmax(M) \odot \mathbf{V}.
\end{equation}

Finally, the attention map $A$ is convolved by a 1x1 convolution and added to the original input $X$:
\begin{equation}
X' = Conv_{1x1}(A) + X.
\end{equation}

The overall complexity of our method is $O(N\log N)$, which significantly reduces the computational cost compared to traditional Window Attention.

Our method maintains the advantages of local window-based attention, such as preserving local structures and reducing memory requirements, while overcoming its limitations by effectively capturing long-range dependencies and lowering computational complexity.

\subsubsection{Dynamic Frequency FeedForward Network}
To address the issue that not all low and high frequency information are beneficial for effective image recovery, we propose an adaptive method through a Frequency Filtering Network (FFN). The key challenge is how to effectively determine which frequency information is crucial. Inspired by the JPEG compression algorithm, we introduce a learnable querying mechanism $Q_{\text{Learned}}$. 
Similar to ~\cite{chi2020fast, rao2021global, kong2023efficient}, this approach employs a gating mechanism between the FFT and IFFT operations for filtering. The method differs in its simplicity, directly using a single self-learned parameter to apply identical filtering across all windows.
The essence of the learnable query lies in a customizable matrix. It essentially acts as a self-learned prompt, influenced by backpropagation, to learn appropriate parameters from the dataset itself for filtering the frequency domain representations of feature maps.
The steps of the proposed method are represented as follows:
\begin{algorithm}
\caption{Frequency Filtering Network Process}
\begin{algorithmic}[1]
\State $I' \gets \Phi(I)$ \Comment{1x1 Convolution to increase channels}
\State $W_{\text{original}} \gets \omega(I')$ \Comment{Window partition}
\State $W_{\text{freq}} \gets \mathcal{F}(W_{\text{original}})$ \Comment{FFT to frequency domain}
\State $W_{\text{filtered}} \gets W_{\text{freq}} \odot Q_{\text{Learned}}$ \Comment{Hadamard product}
\State $W_{\text{spatial}} \gets \mathcal{F}^{-1}(W_{\text{filtered}})$ \Comment{IFFT to spatial domain}
\State $I'' \gets \rho(W_{\text{spatial}})$ \Comment{Restoration of image from windows}
\State $O \gets \Phi^{-1}(I'')$ \Comment{1x1 Convolution to decrease channels}
\end{algorithmic}
\end{algorithm}

Here, $\Phi$ represents the 1x1 convolution operation for expanding the number of channels, $\omega$ stands for the operation of dividing the image into windows, $\mathcal{F}$ denotes the Fast Fourier Transform (FFT), $\mathcal{F}^{-1}$ is the Inverse Fast Fourier Transform (IFFT), $\rho$ stands for the restoration of windows back into a single image, and $\Phi^{-1}$ signifies the 1x1 convolution operation for reducing the number of channels. By employing this sequence of operations, our method ensures that only the necessary frequency information is preserved, leading to a more effective image recovery.

\subsection{Image Fusion Block}
Our approach adopts a convolutional pathway, transforming the input source image into a distinctive feature representation using convolutional layers. This is mathematically represented as follows:
\begin{equation}
F = Conv(\mathbf{I}),
\end{equation}
where $F$ is the resultant feature representation, $Conv$ represents the multiple convolution operations, and $\mathbf{I}$ is the input source image. Subsequently, an amalgamation of Max Pooling and Average Pooling is employed for downsampling. The outcome of this is concatenated to realize a perception at various scales.
\begin{equation}
F' = Concat(AvgPool(F), MaxPool(F)),
\end{equation}
where $F'$ is the downscaled feature representation. During the processing of low-scale feature maps, we address feature displacement using skip connections during the upsampling phase. The corrected features are then convolved to generate an attention map,
\begin{equation}
A = Conv(Connect(UpSample(F'), F)).
\end{equation}

In the final stage, the attention map is used to generate two images, which are element-wise multiplied with the corresponding image. The summation of these results forms the initial fused image, 
\begin{equation}
Fused_{initial} = \sum_{i=1}^{2} A_{i} \odot I_{i}.
\end{equation}

In this scheme, we are able to generate an image, imbued with complementary information extracted from the source image. Nevertheless, it's critical to note that while the attention mechanism retains and integrates a wealth of information from different exposure levels, it does not account for color calibration and exposure correction. Therefore, it is imperative to execute subsequent processing on the initial fused image.

\subsection{Loss Function}
In our proposed method, two types of losses have been incorporated, each contributing to the overall loss function proportionally with respect to their weights. The overall loss function is as follows:
\begin{equation}
\mathcal{L}_{total} = \lambda_{1} \mathcal{L}_{mse} + \lambda_{2} \mathcal{L}_{pyr}.
\end{equation}

\textbf{Reconstruction Loss.} The first component of the overall loss function is the reconstruction loss. The aim of the reconstruction loss is to measure the dissimilarity between the ground truth image and the image reconstructed by our method. We employ the Mean Squared Error (MSE) as our metric for this. The formula for the reconstruction loss is defined as follows:
\begin{equation}
\mathcal{L}_{mse}=\sum_{p=1}^{3 h w}\left|\mathbf{O}^{1}(p)-\mathbf{G}(p)\right|,
\end{equation}
where $\mathbf{O}^{1}$ is the final output image corrected by our method, and $\mathbf{G}$ is the ground truth image. Here, $p$ denotes the pixel index, with $h$ and $w$ being the height and width of the output image, respectively.

\textbf{Laplacian Pyramid Loss.} The second component of our overall loss function is the Laplacian pyramid loss. This is used to measure the dissimilarity between the ground truth image and the image produced by our method on different levels of the Gaussian pyramid. The formula for the Laplacian pyramid loss is:
\begin{equation}
\mathcal{L}_{pyr}=\sum_{i=2}^{n} 2^{(i-2)} \sum_{p=1}^{3 h_{i} w_{i}}\left|\mathbf{O}^{i}(p)-\mathbf{G}^{i}(p)\right|,
\end{equation}
where $\mathbf{O}^{i}$ is the output of the restoration stage at the $i$th level of the pyramid, and $\mathbf{G}^{i}$ is the $i$th level of the Gaussian pyramid, formed from the ground truth image. Here, $p$ denotes the pixel index, $h_{i}$ and $w_{i}$ are the height and width of the image at the $i$th level of the pyramid, respectively, and $n$ is the total number of levels in the pyramid.

In the overall loss function, $\lambda_{1}$ and $\lambda_{2}$ are the weights associated with the reconstruction loss and Laplacian pyramid loss, respectively. 

\section{Experiment}
In our research, we endeavor to benchmark our approach against the state-of-the-art methods for exposure correction, low-light enhancement, and multi-exposure fusion, utilizing several classic datasets for both quantitative and qualitative comparison. Our method's applicability is further demonstrated by extending it to high-level tasks. In a bid to corroborate the versatility of our proposed method, we utilized our exposure correction technique as a pre-enhancer for low-light images. This preparatory step significantly improved the performance of subsequent high-level tasks such as low-light face detection and low-light semantic segmentation. The performance metrics of these tasks serve to underscore the robustness of our approach in real-world applications, bridging the gap between low-level image enhancement and high-level visual understanding. 
\subsection{Datasets for Exposure-related Tasks}
In our study, we thoroughly evaluated our proposed method across five distinct tasks: exposure correction, low-light enhancement, multi-exposure fusion, low-light face detection, and low-light semantic segmentation. In the following sections, we detail the datasets selected for this evaluation, outlining their key characteristics and the reasons for their inclusion in our experimental setup.

\subsubsection{Datasets for Exposure Correction} 
Our experiments involve two exposure-errors datasets, MSEC and LCDP. MSEC dataset contains 24,330 8-bit sRGB images divided into 17,675 training images, 750 validation images, and 5905 test images. The images in them are tuned by the MIT-Adobe FiveK dataset with 5 different exposure values (EV) ranging from under-exposure to over-exposure conditions. Each image of the training set is accompanied by a ground truth image. Each image of the test set has manual correction results from 5 different experts (A/B/C/D/E). The LCDP has 1733 pairs of images, which are divided into 1415 pairs for training, 100 pairs for validation, and 218 pairs for testing. 
Each image in the MSEC dataset has an overall exposure error, while each image in LCDP has different types of exposure errors in different regions. These two datasets represent the types of exposure errors commonly found in reality and are the only two large-scale datasets available for the exposure correction.

\subsubsection{Datasets for Low-light Enhancement}

The LOL dataset, introduced by Wei et al., is specifically designed for low-light image enhancement. It comprises 500 image pairs, each consisting of a low-light image and its corresponding normally lit image. The images in the dataset are categorized into two types: 400 image pairs are selected from the web and serve as the training set, while the remaining 100 pairs, captured by various smartphones in real-life low-light scenarios, form the test set.
The LOL dataset provides a challenging and practical environment for low-light enhancement algorithms, as it covers a diverse range of scenarios, such as indoor, outdoor, dawn, night and backlit scenes, effectively mimicking real-world conditions.

The MIT-Adobe FiveK dataset, introduced by Bychkovsky et al., is a large-scale dataset originally designed for color enhancement and editing. It consists of 5000 high-quality RAW photographs, each retouched by five different photographers, providing a total of 25000 enhanced images.
For the purpose of low-light enhancement, a subset of this dataset is commonly used. This subset includes images captured under various challenging lighting conditions, such as at sunset, under cloudy weather, or indoors with artificial lighting. Each image in this subset is accompanied by multiple retouched versions, offering multiple possible 'ground truths' and thus promoting a more comprehensive evaluation of the enhancement methods.

\subsubsection{Datasets for Multi-eposure Fusion}
In our research, we cultivated a specific subset of 490 image sequences from the SICE dataset, each sequence consisting of an over-exposed, under-exposed, and a high-quality reference image, handpicked to represent extreme exposure scenarios. A set of 360 sequences was randomly chosen for training, with the remaining 130 sequences utilized for validation. For experimental comparison, we adopted 100 randomly picked image pairs from the SICE dataset, supplemented with an additional 18 pairs without ground truth, thus gauging the versatility of our method under various conditions.

\subsubsection{Baseline and Datasets for High-Level Tasks}
We utilized the S3FD\cite{zhang2017s3fd}, a well-known face detection algorithm to evaluate the dark face detection performance and adopted the PSPNet\cite{zhao2017pyramid} as the baseline to evaluate the segmentation performance.Low-light human face detection and low-light semantic segmentation are experimented on the DARKFACE \cite{yang2020advancing} and ACDC \cite{sakaridis2021acdc} datasets respectively. To enhance adaptability, we fine-tuned all visual models on the corrected output.

The DARKFACE dataset \cite{yang2020advancing} is designed for low-light human face detection. It includes 6,000 low-light images from various real-world settings, labeled with bounding boxes identifying human faces. Additionally, the dataset consists of 9,000 unlabeled low-light images and a unique subset of 789 images captured under both low-light and normal lighting conditions. A hold-out testing set of 4,000 low-light images, annotated with human face bounding boxes, is also provided.

The ACDC dataset \cite{sakaridis2021acdc} is used for semantic segmentation under adverse conditions. It comprises 4,006 images, equally distributed under four common adverse conditions: fog, nighttime, rain, and snow. Each image in the dataset is accompanied by fine pixel-level semantic annotations and a binary mask. This mask differentiates between regions of clear and uncertain semantic content within the image, supporting both standard semantic segmentation and uncertainty-aware semantic segmentation. 

\subsection{Metrics for Image Quality Assessment}
Image quality assessment is a fundamental aspect of various processes in computer vision and image processing. The reliability and effectiveness of the proposed methodologies are quantitatively evaluated based on several key metrics. This paper particularly leverages the Peak Signal to Noise Ratio (PSNR), Structural Similarity Index Measure (SSIM) and Multi-Exposure Fusion Structural Similarity Index (MEF-SSIM) to gauge the quality of processed images and thereby ascertain the performance of the proposed approach.
\subsubsection{Peak Signal to Noise Ratio (PSNR)}
The Peak Signal to Noise Ratio (PSNR) is a commonly used objective metric for assessing the quality of reconstruction of lossy compression codecs for image and video data. It is a simple yet effective measure of the error between a reference image and a distorted version of the image.

The PSNR is formally defined as:
\begin{equation}
\mathrm{PSNR} = 10 \cdot \log_{10} \left(\frac{\mathrm{MAX_{I}^{2}}}{\mathrm{MSE}}\right),
\end{equation}
where $\mathrm{MAX_{I}}$ is the maximum possible pixel value of the image. For an 8-bit grayscale image, the maximum pixel value is 255. $\mathrm{MSE}$ is the Mean Squared Error, which is the average squared difference between the pixels of the reference image and the distorted image.

\subsubsection{Structural Similarity Index Measure (SSIM)}

The Structural Similarity Index Measure (SSIM) is a perception-based model that considers changes in structural information, illumination, and contrast as separate components that contribute to the quality of an image.

The SSIM index is calculated as:
\begin{equation}
\mathrm{SSIM(\mathbf{x},\mathbf{y})} = \frac{(2\mu_x\mu_y + c_1)(2\sigma_{xy} + c_2)}{(\mu_x^2 + \mu_y^2 + c_1)(\sigma_x^2 + \sigma_y^2 + c_2)},
\end{equation}
where $\mu_x$ and $\mu_y$ are the average of $\mathbf{x}$ and $\mathbf{y}$; $\sigma_x^2$ and $\sigma_y^2$ are the variance of $x$ and $y$; $\sigma_{xy}$ is the covariance of $x$ and $y$; $c_1=(k_1L)^2$, $c_2=(k_2L)^2$ are two variables to stabilize the division with weak denominator; $L$ is the dynamic range of pixel-values; $k_1=0.01$ and $k_2=0.03$ by default.

\subsubsection{Multi-Exposure Fusion Structural Similarity Index Measure (MEF-SSIM)}

The Multi-Exposure Fusion Structural Similarity Index is an extension of the traditional SSIM that is specifically designed for evaluating the quality of images fused from multiple exposures. It takes into account the quality of the details in the fused image, the naturalness of the fused image, and the visibility of the image content in different exposures.

The MEF-SSIM index is calculated as:
\begin{equation}
\mathrm{MEF-SSIM(\mathbf{I_{f}, I_{i}})} = \mathrm{SSIM(\mathbf{I_{f}, I_{i}})} \cdot W_{i},
\end{equation}
where $\mathbf{I_{f}}$ is the fused image and $\mathbf{I_{i}}$ is the $i$-th source image. $W_{i}$ is the weight of the $i$-th source image, calculated as:
\begin{equation}
W_{i} = \frac{\exp(-\beta \cdot (\mathbf{I_{i}} - \bar{I})^{2})}{\sum_{j=1}^{n}\exp(-\beta \cdot (\mathbf{I_{j}} - \bar{I})^{2})},
\end{equation}
where $\bar{I}$ is the mean intensity of the $i$-th source image, $\beta$ is a parameter controlling the strength of the weighting function, and $n$ is the number of source images. 
By weighing the contribution of each source image to the final fused image based on its similarity to the fused image and its visibility, the MEF-SSIM provides a more accurate and robust measure of the quality of multi-exposure fusion images.

\begin{figure*}[!htb]
\centering
\setlength{\tabcolsep}{1pt}
\resizebox{0.9\textwidth}{!}{
\begin{tabular}{c}
\includegraphics{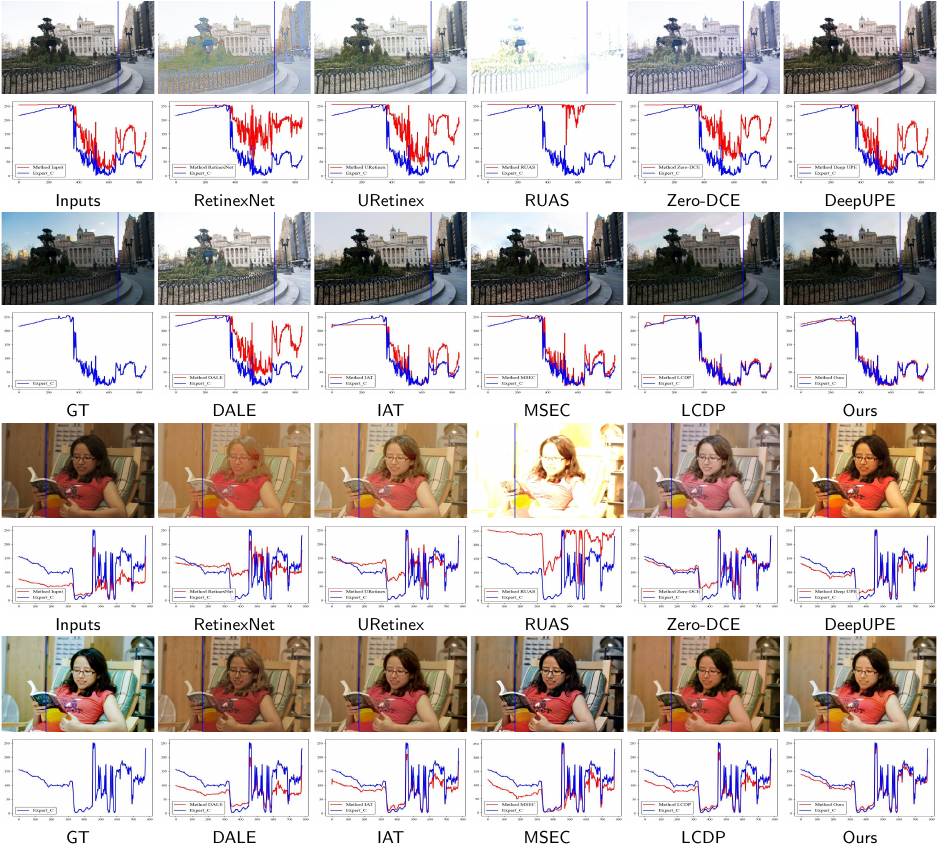}
\end{tabular}}
\caption{Qualitative Comparison of exposure correction performance on MSEC dataset.}
\label{fig:msec}
\end{figure*}

\begin{figure*}[!htb]
\centering 
\setlength{\tabcolsep}{1pt}
\resizebox{0.9\textwidth}{!}{
\begin{tabular}{c}
\includegraphics{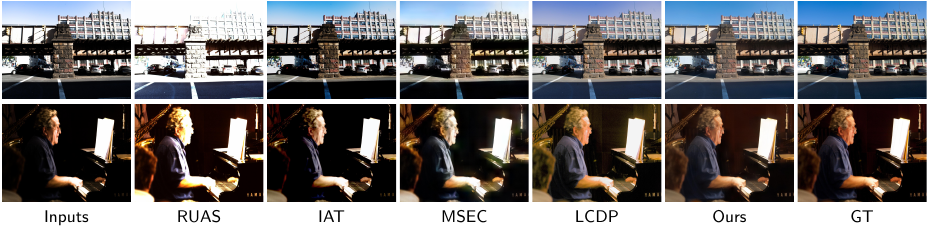}
\end{tabular}}
\caption{Qualitative Comparison of exposure correction performance on LCDP dataset.}
\label{fig:lcdp}
\end{figure*}

\subsection{Parameter Settings}
All of our experiments were run on one NVIDIA GeForce RTX 4090 GPU. The optimizer chose ADAM, with a learning rate of 2e-4. The weight of the loss function is set to $\lambda_{1} = 1.0 $ and $\lambda_{2} = 1.0 $. The level of the Laplacian pyramid decomposition is $N=4$. Before training, we process the data set and scale the image to $512\times512$ size before inputting it into the network for training. Guided by the JPEG compression methodology, we empirically establish a patch size of 8 × 8 for both the weight matrix estimation and the computation of self-attention, ensuring consistency and uniformity across our calculations.

\begin{table*}[t]
\centering
\caption{Quantitative comparison of exposure correction performance on MSEC dataset. }
\resizebox{1\textwidth}{!}{
\begin{tabular}{l||cc||cc||cc||cc||cc||cc}
\toprule
\multicolumn{1}{c}{}                         & \multicolumn{2}{c}{Expert A}                                                                        & \multicolumn{2}{c}{Expert B}                                                                        & \multicolumn{2}{c}{Expert C}                                                                        & \multicolumn{2}{c}{Expert D}                                                                        & \multicolumn{2}{c}{Expert E}                                                                        & \multicolumn{2}{c}{Avg}   \\ \cmidrule(lr){2-3}\cmidrule(lr){4-5}\cmidrule(lr){6-7} \cmidrule(lr){8-9}\cmidrule(lr){10-11} \cmidrule(lr){12-13}
\multicolumn{1}{c}{\multirow{-2}{*}{Method}} & \multicolumn{1}{c}{\cellcolor[HTML]{EFEFEF}PSNR} & \multicolumn{1}{c}{\cellcolor[HTML]{EFEFEF}SSIM} & \multicolumn{1}{c}{\cellcolor[HTML]{EFEFEF}PSNR} & \multicolumn{1}{c}{\cellcolor[HTML]{EFEFEF}SSIM} & \multicolumn{1}{c}{\cellcolor[HTML]{EFEFEF}PSNR} & \multicolumn{1}{c}{\cellcolor[HTML]{EFEFEF}SSIM} & \multicolumn{1}{c}{\cellcolor[HTML]{EFEFEF}PSNR} & \multicolumn{1}{c}{\cellcolor[HTML]{EFEFEF}SSIM} & \multicolumn{1}{c}{\cellcolor[HTML]{EFEFEF}PSNR} & \multicolumn{1}{c}{\cellcolor[HTML]{EFEFEF}SSIM} & \multicolumn{1}{c}{\cellcolor[HTML]{EFEFEF}PSNR} & \multicolumn{1}{c}{\cellcolor[HTML]{EFEFEF}SSIM} \\ \midrule
WVM                     & 14.488          & 0.788          & 15.803          & 0.699          & 15.117          & 0.678          & 15.863          & 0.693          & 16.469          & 0.704          & 15.548          & 0.688          \\
LIME                    & 11.154          & 0.591          & 11.828          & 0.610          & 11.517          & 0.607          & 12.638          & 0.628          & 13.613          & 0.653          & 12.150          & 0.618          \\
HDR CNN w/PS            & 15.812          & 0.667          & 16.970          & 0.699          & 16.428          & 0.681          & 17.301          & 0.687          & 18.650          & 0.702          & 17.032          & 0.687          \\
DPED (iPhone)           & 15.134          & 0.609          & 16.505          & 0.636          & 15.907          & 0.622          & 16.571          & 0.627          & 17.251          & 0.649          & 16.274          & 0.629          \\
DPED (BlackBerry)       & 16.910          & 0.642          & 18.649          & 0.713          & 17.606          & 0.653          & 18.070          & 0.679          & 18.217          & 0.668          & 17.890          & 0.671          \\
DPE (HDR)               & 15.690          & 0.614          & 16.548          & 0.626          & 16.305          & 0.626          & 16.147          & 0.615          & 16.341          & 0.633          & 16.206          & 0.623          \\
DPE (S-FiveK)           & 16.933          & 0.678          & 17.701          & 0.668          & 17.741          & 0.696          & 17.572          & 0.674          & 17.601          & 0.670          & 17.510          & 0.677          \\
RetinexNet              & 10.759          & 0.585          & 11.613          & 0.596          & 11.135          & 0.605          & 11.987          & 0.615          & 12.671          & 0.636          & 11.633          & 0.607          \\
Deep UPE                & 13.161          & 0.610          & 13.901          & 0.642          & 13.689          & 0.632          & 14.806          & 0.649          & 15.678          & 0.667          & 14.247          & 0.640          \\
Zero-DCE                & 11.643          & 0.536          & 12.555          & 0.539          & 12.058          & 0.544          & 12.964          & 0.548          & 13.769          & 0.580          & 12.597          & 0.549          \\
RUAS & 10.166 & 0.391 & 10.522 & 0.440 & 9.356 & 0.411 & 11.013 & 0.441 & 11.574 & 0.466 & 10.526 & 0.430 \\
URetinex                & 11.420          & 0.632          & 12.230          & 0.700          & 11.818          & 0.672          & 13.078          & 0.701          & 14.066          & 0.735          & 12.522          & 0.688          \\
DALE & 13.294 & 0.691 & 14.324 & 0.757 & 13.734 & 0.722 & 14.256 & 0.743 & 14.511 & 0.763 & 14.024 & 0.735 \\
IAT(local)              & 16.610          & 0.750          & 17.520          & 0.822          & 16.950          & 0.780          & 17.020          & 0.780          & 16.430          & 0.789          & 16.910          & 0.783          \\
MSEC                    & 19.158          & 0.746          & 20.096          & 0.734          & 20.205          & 0.769          & 18.975          & 0.719          & 18.983          & 0.727          & 19.483          & 0.739          \\
IAT                     & 19.900          & 0.817          & 21.650          & 0.867          & 21.230          & 0.850          & 19.860          & 0.844          & 19.340          & 0.840          & 20.340          & 0.844          \\
LCDPNet                 & 20.574          & 0.809          & 21.804          & 0.865          & 22.295          & 0.855          & 20.108          & 0.824          & 19.281          & 0.822          & 20.812          & 0.835          \\ 
\textbf{Ours}           & \textbf{20.795} & \textbf{0.821} & \textbf{21.902} & \textbf{0.874} & \textbf{22.812} & \textbf{0.859} & \textbf{20.113} & \textbf{0.837} & \textbf{19.979} & \textbf{0.836} & \textbf{21.120} & \textbf{0.845} \\ \bottomrule
\end{tabular}}
\label{res:msec}
\end{table*}

\begin{table*}[t]
\centering
\caption{Quantitative comparison of exposure correction performance on LCDP dataset.}
\resizebox{1\textwidth}{!}{
\begin{tabular}{lccccccccccc}
\toprule
Method & ZeroDCE & HE & RetinexNet & ClAHE & LIME & MSEC & IAT & DeepUPE & HDRnet & LCDP & \textbf{Ours} \\
\midrule
PSNR & 12.587 & 15.975 & 16.201 & 16.327 & 17.335 & 17.066 & 17.842 & 20.970 & 21.834 & 23.239 & \textbf{23.415} \\
SSIM & 0.653 & 0.684 & 0.631 & 0.642 & 0.686 & 0.642 & 0.684 & 0.818 & 0.818 & 0.842 & \textbf{0.851} \\
\bottomrule
\end{tabular}}
\label{res:lcdp}
\end{table*}

\subsection{Comparison on Exposure Correction}
Our proposed method undergoes evaluation on two classic large-scale datasets, with the selection of cutting-edge techniques for comparison. For the MSEC dataset, we compare with methods including: WVM \cite{fu2016weighted}, LIME \cite{guo2016lime}, HDR CNN \cite{eilertsen2017hdr}, DPED \cite{ignatov2017dslr}, DPE \cite{chen2018deep}, RetinexNet \cite{wei2018deep}, Deep UPE \cite{wang2019underexposed}, Zero-DCE \cite{guo2020zero}, RUAS \cite{liu2021retinex}, URetinex \cite{xu2020star}, DALE \cite{kwon2020dale}, IAT \cite{cui2022illumination}, MSEC \cite{afifi2021learning}, and LCDP \cite{wang2022local}. For the LCDP dataset, we compare with methods including: Zero-DCE \cite{guo2020zero}, HE \cite{pizer1987adaptive}, RetinexNet \cite{wei2018deep}, CLAME \cite{reza2004realization}, LIME \cite{guo2016lime}, MSEC \cite{afifi2021learning}, IAT \cite{cui2022illumination}, Deep UPE \cite{wang2019underexposed}, HDRnet \cite{gharbi2017deep}, and LCDP \cite{wang2022local}.

\subsubsection{Qualitative Comparison}
Figure \ref{fig:msec} and Figure \ref{fig:lcdp} respectively display qualitative comparisons on the MSEC and LCDP datasets. Upon an overall observation, images corrected by our method demonstrate the closest details and colors when compared to reference images. In addition, we verify the results through an intensity signal analysis. For underexposed inputs, Zero-DCE and Deep UPE achieve an overall similarity with the reference image signals, despite local discrepancies. Other methods reveal a large overall deviation from the reference images due to failed global exposure adjustment. For overexposed inputs, IAT, MSEC, and LCDP effectively reduce exposure values but struggle to restore the lost colors, leading to considerable artifact generation. Other methods fail in correctly correcting overexposed images. Our proposed method exhibits the highest accuracy in pixel intensity, closely aligning with the ground truth values.

\subsubsection{Quantitative Comparison}
We employ PSNR and SSIM as the metrics to measure the quality of corrected images. As Table \ref{res:msec} shows, for the MSEC dataset, we compare our results with those from five expert photographers. This comparison method considers the variations in camera-based rendering settings, where professionals might render the same image differently. Therefore, our paper evaluates the proposed method against five expert-rendered images, all of which represent satisfactory exposure reference images. Our method achieves the highest scores on both PSNR and SSIM across all five expert reference sets. These results indicate that the proposed method effectively corrects both overexposed and underexposed input images, delivering high-quality results with accurate pixel intensity distributions and true-to-life textures. 
Table \ref{res:lcdp} presents the results for the LCDP dataset, revealing that our proposed method similarly achieves optimal results. This performance demonstrates that our method not only corrects images with global exposure errors but also handles scenarios with different types of exposure errors in different regions. The superior results across both datasets powerfully exhibit the excellent performance of our method in dealing with exposure-error datasets.

\begin{figure*}[t]
\centering 
\setlength{\tabcolsep}{1pt}
\resizebox{\textwidth}{!}{
\begin{tabular}{c}
\includegraphics{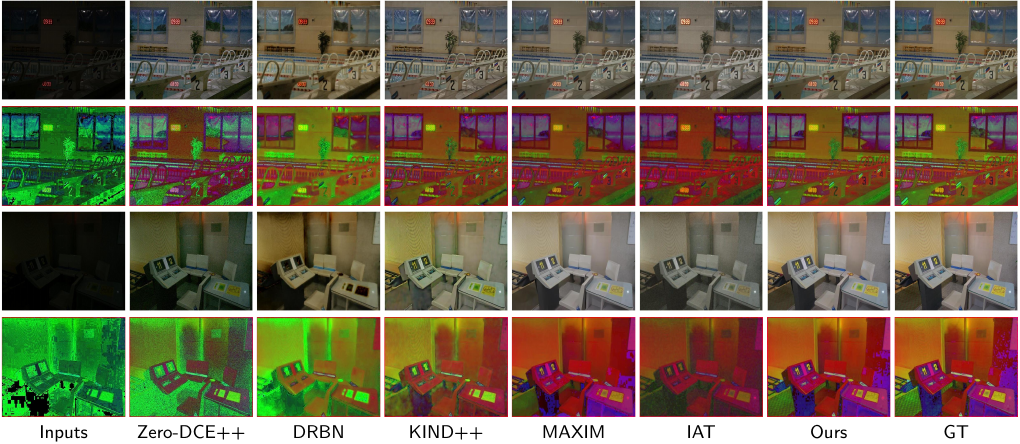}
\end{tabular}}
\caption{Qualitative comparison of low-light enhancement performance on LOL dataset. The RGB image is converted to HSV color mode at the bottom, so that it is easier to compare the details.}
\label{fig:hsv}
\end{figure*}

\begin{figure*}[t]
\centering
\setlength{\tabcolsep}{1pt}
\resizebox{0.9\textwidth}{!}{
\begin{tabular}{cc}
\includegraphics[width=0.43\textwidth,height=0.23\textheight]{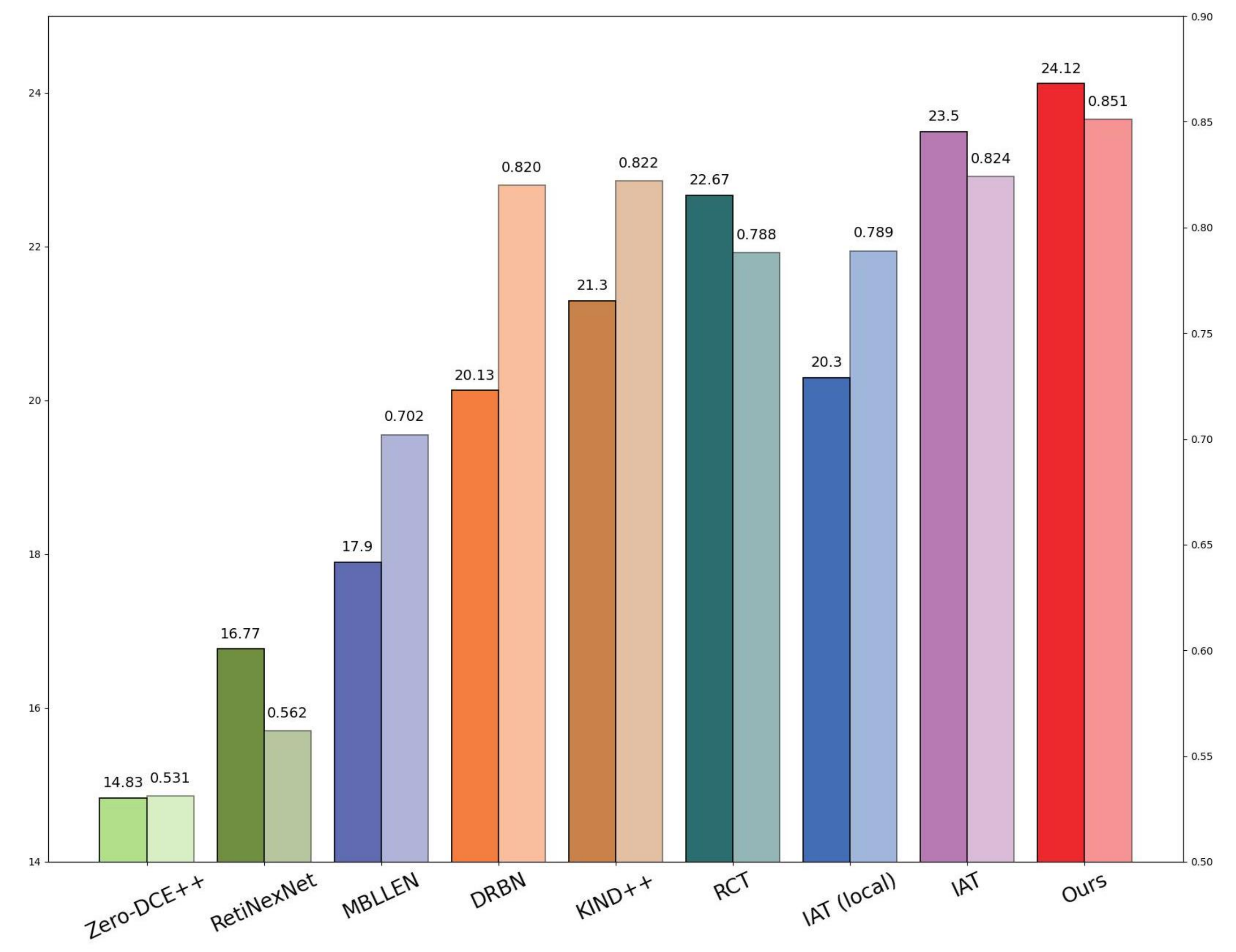}
&\includegraphics[width=0.43\textwidth,height=0.23\textheight]{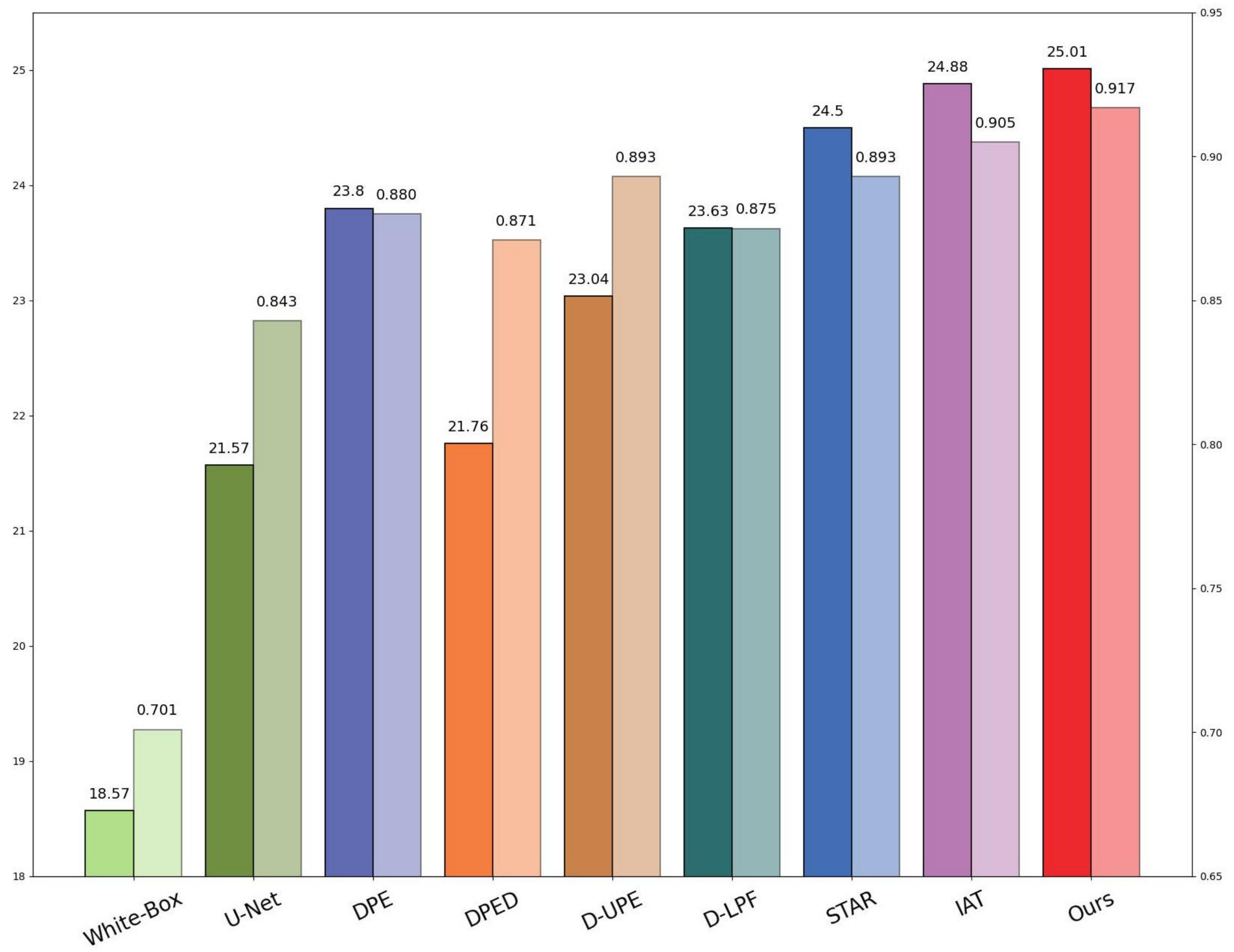}
\\
(a) PSNR and SSIM for LOL & (b) PSNR and SSIM for MIT-Adobe FiveK
\\
\end{tabular}}
\caption{Quantitative comparison of low-light enhancement performance on LOL and MIT-Adobe FiveK dataset. }
\label{res:lol5k}
\end{figure*}

\subsection{Comparison on Low-light Enhancement}
To verify the single exposure correction capability of our method, we not only test its exposure correction performance but also its performance in low-light enhancement tasks.
For the LOL dataset \cite{wei2018deep}, we compare our method with several state-of-the-art methods including Zero-DCE++ \cite{guo2020zero}, RetinexNet \cite{wei2018deep}, MBLLEN \cite{lv2018mbllen}, DRBN \cite{yang2020fidelity}, KIND++ \cite{zhang2021beyond}, RCT \cite{kim2021representative}, and IAT \cite{cui2022illumination}.
For the MIT-Adobe FiveK dataset \cite{bychkovsky2011learning}, we compare our method with White-Box \cite{hu2018exposure}, U-Net \cite{ronneberger2015u}, DPE \cite{chen2018deep}, DPED \cite{ignatov2017dslr}, D-UPE \cite{wang2019underexposed}, D-LPF \cite{moran2020deeplpf}, STAR \cite{zhang2021star}, and IAT \cite{cui2022illumination}.

\subsubsection{Qualitative Comparison}
A key part of our comparison was a detailed qualitative analysis. To better discern the differences between our method and others, we converted all image enhancement results to the HSV color space. This transformation made the disparities in performance more pronounced and visually understandable. The color mappings of the results are presented in Figure \ref{fig:hsv}. In this representation, more similar colors after the mapping operation suggest more consistent brightness and color across corresponding areas.
Through this process, we found that popular methods such as Zero-DCE++, DRBN, and KIND++ failed to holistically enhance the image, resulting in an overall darker image. Other methods we compared showed larger color discrepancies in certain areas when compared to the ground truth (GT), indicating an inherent limitation in capturing global dependencies, thus failing to enhance challenging, less noticeable dark parts.
In stark contrast, our method showcased its ability to enhance both global and local brightness to suitable levels, producing images that not only exhibit accurate brightness levels but also preserve the best details and colors. We have shown these successful enhancements in various sample sets - the diving platform in the first set, the control panel in the second set, and the box in the third set - attesting to the robustness of our method across different scenarios.

\subsubsection{Quantitative Comparison}
We visualize the PSNR and SSIM test results of the two datasets using a dual bar chart, as shown in Figure \ref{res:lol5k}.
On the LOL dataset, our method was unmatched, achieving the highest scores in both PSNR (24.12) and SSIM (0.851), surpassing all other methods in the test. In fact, our method improved upon the second-best PSNR score (IAT's 23.50) by a significant margin of 2.6\% and topped the second-best SSIM score (IAT's 0.824) by an impressive 3.3\%.
The same superiority of our method was observed on the MIT-Adobe FiveK dataset, where our method again delivered the best performance, achieving a PSNR score of 25.01 and an SSIM score of 0.917. Compared to the second-best method (IAT), our method enhanced the PSNR and SSIM scores by 0.5\% and 1.3\%, respectively. 
These results highlight the robustness and consistency of our method in enhancing images with minimal loss of structural and textural information. The comprehensive experimental comparison we have conducted showcases the outstanding performance of our method in low-light enhancement.

\begin{table*}[t]
\centering
\caption{Quantitative comparison of multi-exposure fusion performance on SICE dataset.}
\resizebox{0.98\textwidth}{!}{
\begin{tabular}{lcccccccccccccc}
\toprule
Metric & MGFF & PMGI & MEFCNN & MEFCL & DeepFuse & DSIFT & U2Fusion & IFCNN & MEFGAN & AGAL & CFNET & DPEMEF  & \textbf{Ours} \\
\midrule
PSNR & 19.19 & 17.42 & 14.19 & 19.32 & 17.58 & 15.38 & 17.67 & 19.13 & 19.71 & 19.91 & 20.35 & 19.23  & \textbf{21.45}\\
SSIM & 0.894 & 0.868 & 0.752 & 0.901 & 0.883 & 0.813 & 0.863 & 0.894 & 0.902 & 0.919 & 0.908 & 0.904  & \textbf{0.941}\\
MEF-SSIM & 0.820 & 0.899 & 0.875 & 0.908 & 0.843 & 0.848 & 0.897 & 0.896 & 0.819 & 0.868 & 0.904 & 0.916   & \textbf{0.926}\\
\bottomrule
\end{tabular}}
\label{table:sice}
\end{table*}

\begin{figure*}[!htb]
\centering
\setlength{\tabcolsep}{1pt}
\resizebox{0.95\textwidth}{!}{
\begin{tabular}{c}
\includegraphics{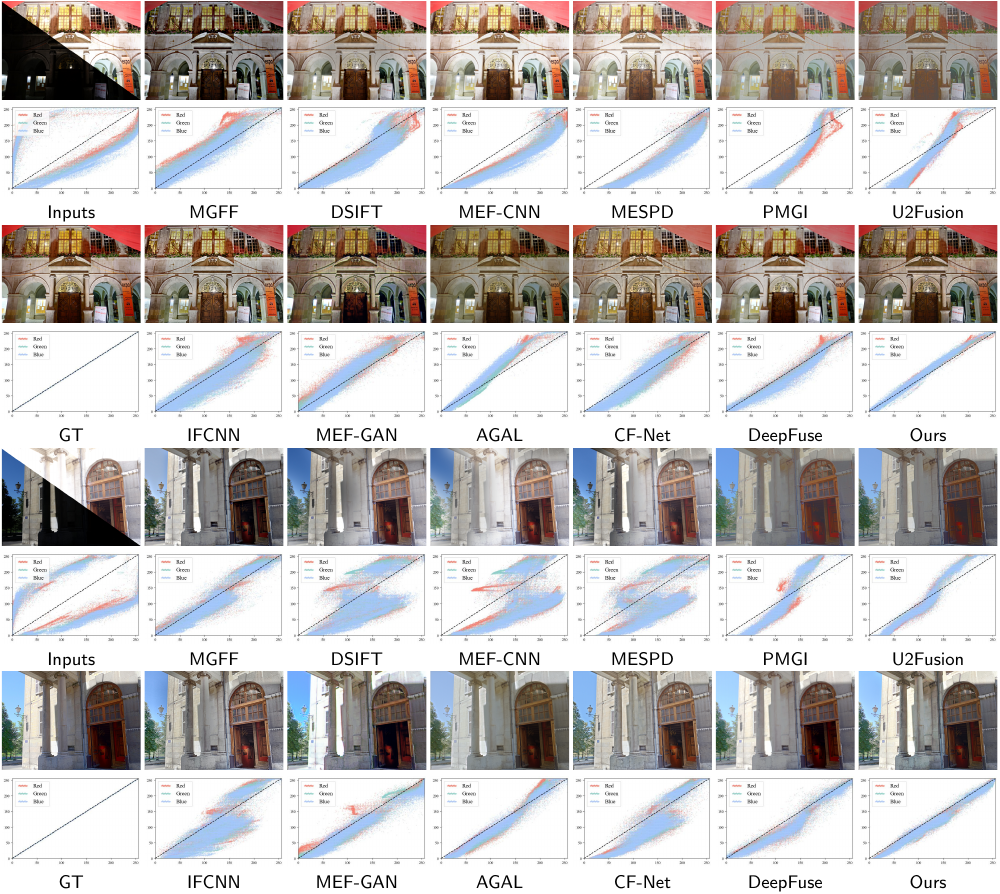}
\end{tabular}}
\caption{ Qualitative comparison of multi-exposure fusion performance on SICE dataset. The signal contrast below the image is derived from the per-channel RGB mapping from the fused image to the GT image. }
\label{fig:mef}
\end{figure*}

\subsection{Comparison on Multi-Exposure Fusion}
Our method is also used in the processing of exposure fusion. It is subjected to a comparative study against eleven leading-edge algorithms. This includes a trio of classic algorithms, specifically, MGFF\cite{GFF}, DSIFT\cite{liu2015dense}, MEF-CNN\cite{MEF-CNN}, and a group of eight advanced algorithms built upon the foundations of deep learning, such as, DeepFuse\cite{Deepfuse}, PMGI\cite{PMGI}, U2Fusion\cite{U2}, MEF-GAN\cite{MEF-GAN}, AGAL\cite{liu2022attention}, CF-NET\cite{CF-NET}, and DPE-MEF\cite{DPE-MEF}.

%

\begin{table*}[t]
\centering
\caption{Performance comparisons on high-level vision tasks. We retrain the detector/segmentator in all cases containing the enhancer.}
\resizebox{0.98\textwidth}{!}{
\begin{tabular}{|c|c|c|c|c|c|c|c|c|c|c|}
\hline
Task & \multicolumn{3}{c|}{Dark Face Detector} & \multicolumn{7}{c|}{Enhancer + Detector (Finetune)}  \\
\hline
Method & HLA          & REG         & MEAT    &LIME   & ZeroDCE & MSEC  & RUAS & LCDP  & IAT   & \textbf{Ours} \\
\hline
mAP & 0.607       & 0.514       & 0.526    &0.644  & 0.665   & 0.659 & 0.642 & 0.654 & 0.663 & \textbf{0.677}     \\
\hline \hline
Task & \multicolumn{3}{c|}{Nighttime Semantic Segmentator} & \multicolumn{7}{c|}{Enhancer + Segmentator (Finetune)} \\
\hline
Method & DANNet         & CIC        & GPS-GLASS     &LIME      & ZeroDCE  & MSEC   & RUAS  & LCDP  & IAT  & \textbf{Ours} \\ \hline
mIoU & 0.398          & 0.264         & 0.380        &0.447     & 0.452    & 0.449  & 0.448 & 0.455 & 0.456 & \textbf{0.468} \\ \hline   
\end{tabular}}
\label{res:high}
\end{table*}

\begin{figure*}[!htb]
\centering
\setlength{\tabcolsep}{1pt}
\begin{tabular}{c}
\includegraphics[width=0.98\textwidth]{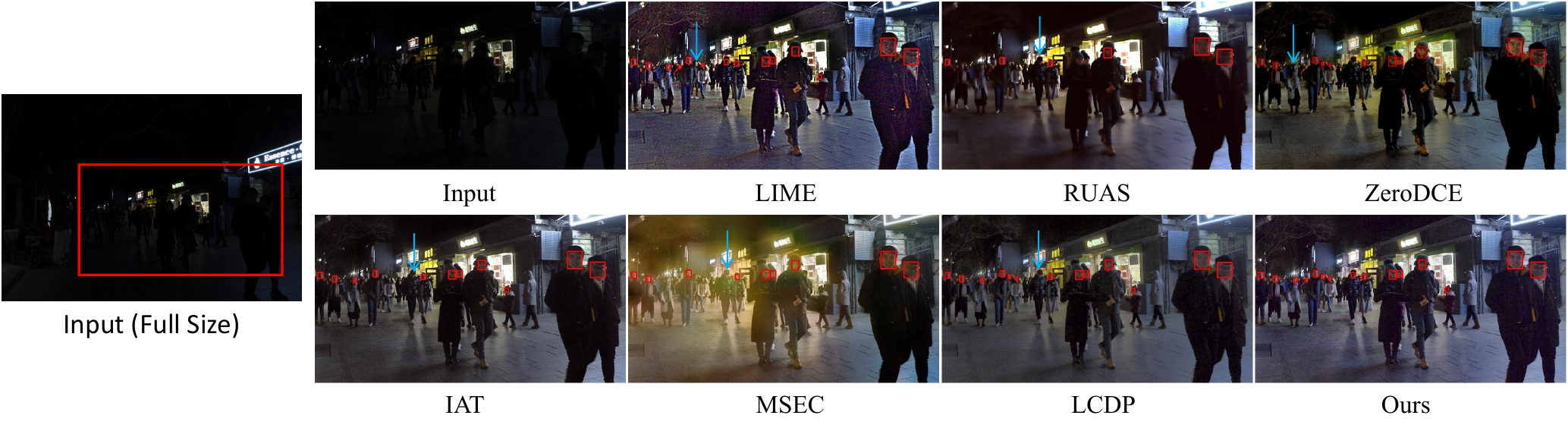} \\
\includegraphics[width=0.98\textwidth]{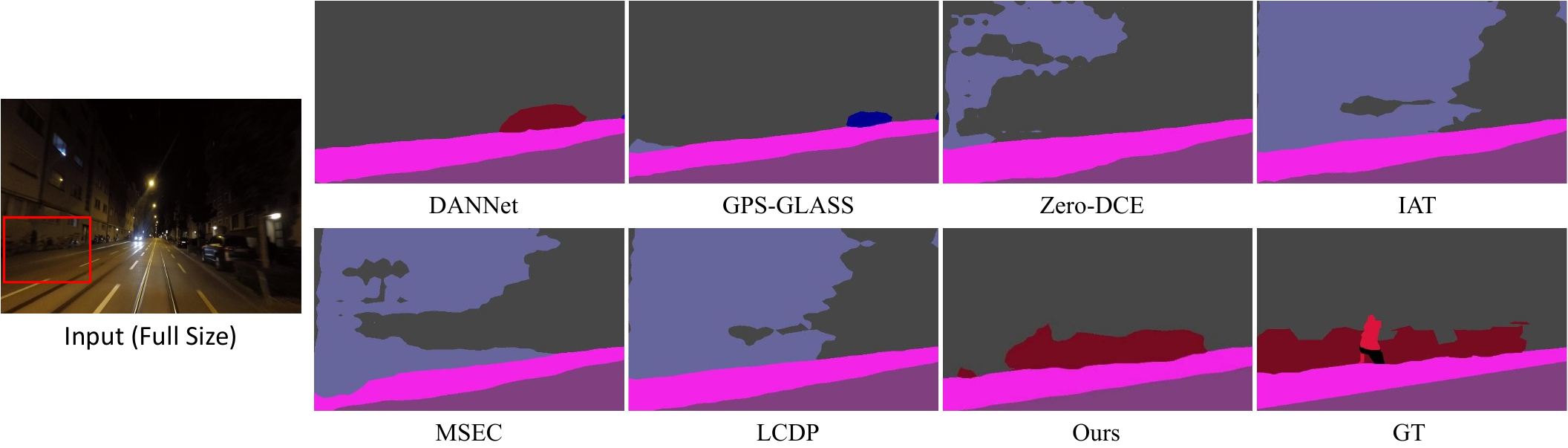}
\end{tabular}
\caption{Qualitative comparison on high-level tasks.}
\label{fig:high}
\end{figure*}

\subsubsection{Qualitative Comparison}
The qualitative comparison with other methods is depicted in Figure \ref{fig:mef}. To discern the difference between each fused method and the Ground Truth (GT) image more clearly, we map each pixel in the RGB channel of each image to the corresponding pixel in the RGB channel of the GT image and depict it beneath each image group. The closer the RGB mapping curve is to the diagonal, the closer each fused image is to the details and colors of the GT image.
The images fused by MGFF, DSIFT, MEF-CNN, MESPD, PMGI, and U2Fusion exhibit significant exposure misalignment and ghosting, as indicated by the large deviation of the mapping curve from the GT. Other comparison methods tend to be linear overall, but suffer from localized detail distortion, resulting in significant bulges in the mapping curve. In contrast, our method can achieve the best colors and details after fusion, and the RGB mapping curve is the smoothest and closest to the diagonal.

\subsubsection{Quantitative Comparison}
Table \ref{table:sice} presents the comparison of our method with others in terms of three key metrics: PSNR, SSIM, and MEF-SSIM. As can be observed, our method significantly outperforms all existing methods specifically designed for multi-exposure fusion. This highlights the powerful generalization ability of our proposed method, which can not only correct the direct input exposure errors, but also reasonably correct the exposure after the fusion of the blocks to generate the image.

\subsection{Extending to Hige-Level Tasks}
To demonstrate the generalizability of our method, we apply it to relevant high-level vision tasks, including low-light face detection and low-light semantic segmentation tasks. To thoroughly evaluate its performance, we contrast it not only with some correction methods but also consider specific detection methods including HLA \cite{wang2022unsupervised}, REG \cite{liang2021recurrent}, MAET \cite{cui2021multitask}, and segmentation methods including DANNet \cite{xu2020star}, CIC \cite{lengyel2021zero}, GPS-GLASS \cite{lee2022gps}. 

\begin{figure*}[!htb]
\centering
\setlength{\tabcolsep}{1pt}
\begin{tabular}{c}
\includegraphics[width=0.95\textwidth]{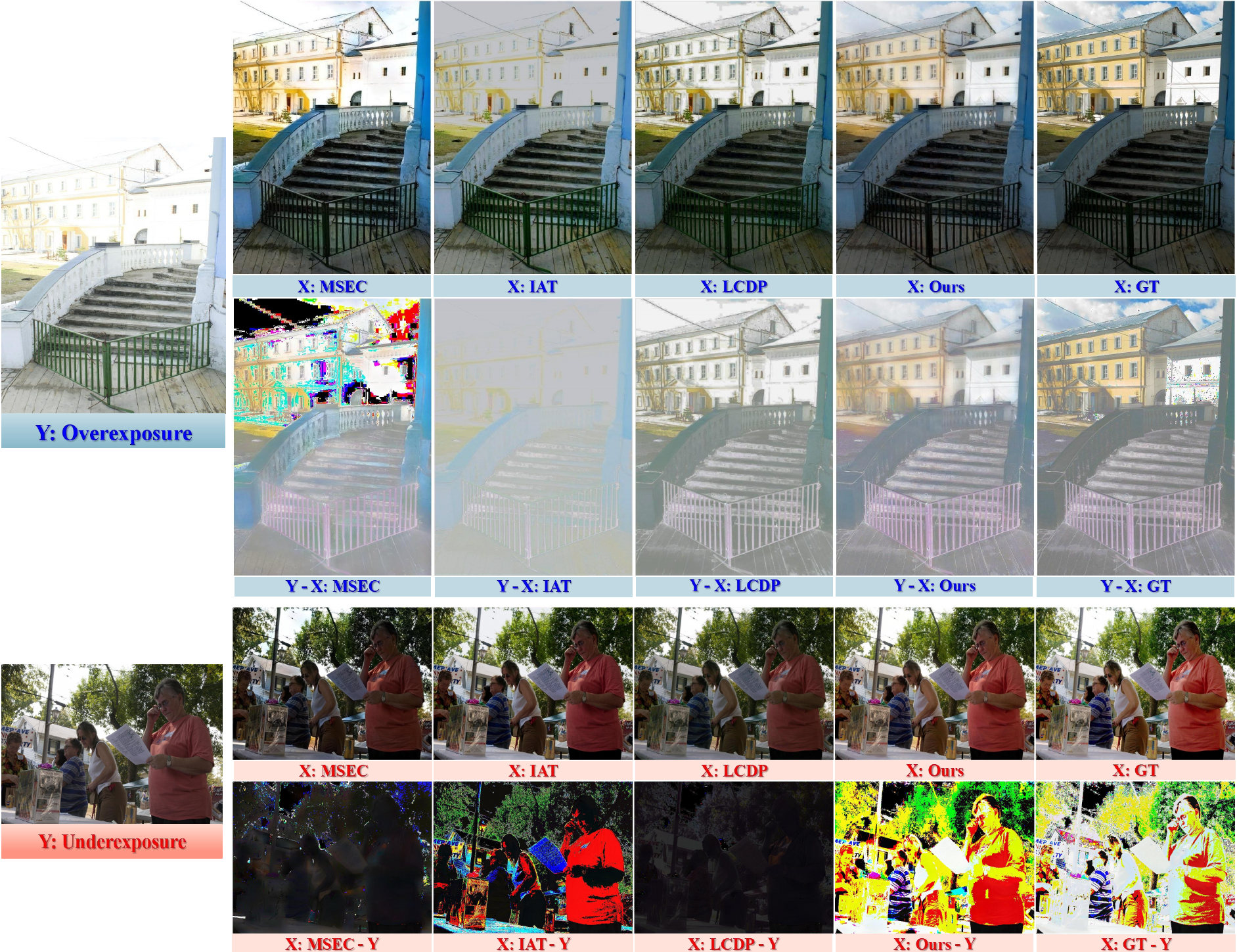}
\end{tabular}
\caption{Visualization of the exposure compensation. The compensation map below the image is obtained by performing absolute subtraction between the corrected image and the GT image.}
\label{fig:gain}
\end{figure*}

\subsubsection{Qualitative Comparison}
We apply our proposed method as an enhancer for low-light images, with the enhanced images input into the baseline network for low-light face detection and low-light semantic segmentation. The comparison results with other methods are shown in Figure \ref{fig:high}. For the results of low-light face detection, images enhanced by other methods noticeably suffer from detail distortion and artifacts. These issues severely impede the subsequent network's detection performance, causing the network to struggle in detecting smaller faces at distance and side faces in close proximity. In contrast, the enhanced images provided by our method achieved the best detection results, effectively identifying these challenging cases. For the results of low-light semantic segmentation, the images enhanced by other methods have limited capabilities for enhancing seriously dark parts, making small-sized objects within difficult to recognize and segment effectively. Our method can effectively restore extremely dark areas for the subsequent network to perform better segmentation.

\subsubsection{Quantitative Comparison}
As shown in Table \ref{res:high}, our method outperforms existing low-light detection and segmentation methods, which do not directly enhance images but proceed with direct detection and segmentation, forming a cascade training network. As the built-in enhancement module is coupled with the detector, it's challenging to further improve. In contrast, the method of enhancing first and detecting later realizes information decoupling, allowing each part to be optimized separately. Under the same baseline network, we outperform all other enhancement methods in both detection and segmentation aspects.

\begin{figure*}[!htb]
\centering
\setlength{\tabcolsep}{1pt}
\resizebox{\textwidth}{!}{
\begin{tabular}{c}
\includegraphics{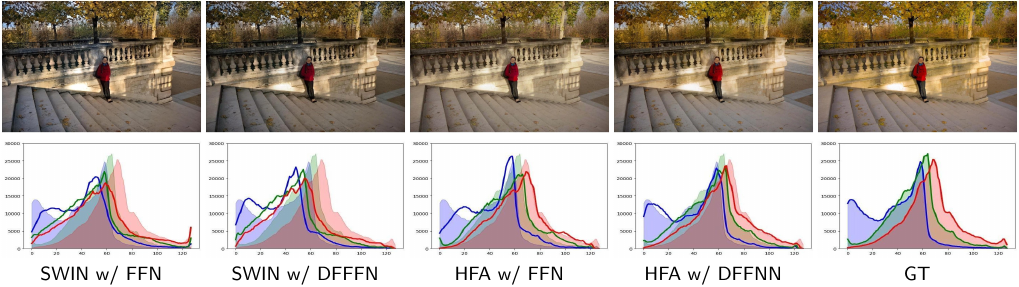}
\end{tabular}}
\caption{Visualization of the proposed module ablation experiment.}
\label{fig:abexp}
\end{figure*}

\begin{table*}[t]
\centering
\caption{Ablation studies on the proposed Holistic Frequency Attention and Dynamic Frequency FeedForward Network.}
\resizebox{0.7\textwidth}{!}{
\begin{tabular}{ccccccc}
\toprule
\multirow{2}{*}{Options} & \multicolumn{2}{c}{Attention} & \multicolumn{2}{c}{Feed-Forward Network} & \multicolumn{2}{c}{Metrics} \\
\cmidrule(lr{0pt}){2-3} \cmidrule(lr{0pt}){4-5} \cmidrule(lr{0pt}){6-7}
                         & HF Attention & SWIN Attention & DF FFN              & FFN                & PSNR          & SSIM        \\
\midrule
SWIN w/ FFN              & \ding{55}    & \ding{51}      & \ding{55}           & \ding{51}          & 22.079        & 0.821       \\
SWIN w/ DFFFN            & \ding{55}    & \ding{51}      & \ding{51}           & \ding{55}          & 22.214        & 0.827       \\
HFA w/ FFN               & \ding{51}    & \ding{55}      & \ding{55}           & \ding{51}          & 22.571        & 0.832       \\
HFA w/ DFFNN             & \ding{51}    & \ding{55}      & \ding{51}           & \ding{55}          & 22.812        & 0.859    
\\ \bottomrule
\end{tabular}}
\label{res:abexp}
\end{table*}

\subsection{Ablation Study}
\subsubsection{Study on Exposure Compensation}
To delve deeper into the specifics of the exposure gain performed by our proposed method when processing input images, we carried out an extensive analysis. Exposure gain, in our case, is accomplished by the absolute difference between the corrected and input images. The visualization of exposure gains for every comparative method is presented in Figure \ref{fig:gain}.

As seen from the figure, our method demonstrates superior performance in achieving optimal exposure gain. It not only accurately identifies every region requiring gain but also applies appropriate enhancement to these areas. Moreover, our approach is capable of increasing details and colors in areas that have already suffered significant distortion.
In comparison, the other methods either fail to provide correct exposure gain, or they are unable to effectively enhance details and colors. This comparative analysis further underlines the effectiveness and precision of our proposed method in terms of exposure gain, especially when dealing with areas that exhibit distortion in details and colors.

\subsubsection{Study on Holistic Frequency Attention}
The spectral domain attention that we propose has significantly reduced the computational complexity. While the original attention's complexity is $O(N^2)$, which is unacceptable for larger input sizes, our method scales at $O(N*\log(N))$, primarily due to the properties of the fast Fourier transform. As illustrated in Table \ref{res:abfast}, the complexity does not increase with the size of the window, unlike window-based methods where the complexity gradually shifts from $O(N)$ to $O(N^2)$ as the window size increases. Our approach remains insensitive to the window size, and neither the FLOPs nor the GPU memory usage increase with the window size.

On the other hand, our approach replaces the spatial matrix multiplication with the multiplication in the frequency domain. This explicit use of frequency domain filtering exhibits superior performance in image restoration tasks when compared to the spatial domain. As Table \ref{res:abexp} shows, replacing Holistic Frequency Attention with Swin Attention in both DF FFN and FFN leads to a significant drop in PSNR and SSIM. The primary reason is that while the shifted window partitioning method reduces the computational cost, it does not fully leverage the useful information between different windows. This verifies that attention based on frequency domain estimation outperforms window-based attention in exposure correction. Furthermore, we visualize the ablation results and show them in Figure \ref{fig:abexp}. It can be observed that the removal of HFA results in exposure imbalance and detail distortion in images.

\begin{table}[h]
\centering
\caption{The computational complexity and test-time costs of frequency domain methods and spatial domain methods.}
\resizebox{1\columnwidth}{!}{
\begin{tabular}{ccccc}
\toprule
               & \multicolumn{2}{c}{Swin Transformer Block} & \multicolumn{2}{c}{Ours HDF Block} \\
\cmidrule(lr{0pt}){2-3} \cmidrule(lr{0pt}){4-5}
Window Size    & FLOPs                & GPU Memory          & FLOPs         & GPU Memory         \\
\midrule
8 $\times$ 8     & 37.8G                & 6.8GB               & 31.3G         & 5.7GB              \\
32 $\times$ 32   & 41.7G                & 11.5GB              & 31.2G         & 5.6GB              \\
64 $\times$ 64   & Out of memory        & Out of memory       & 30.9G         & 5.3GB              \\
128 $\times$ 128 & Out of memory        & Out of memory       & 30.8G         & 5.2GB              \\
512 $\times$ 512 & Out of memory        & Out of memory       & 29.9G         & 5.2GB            
\\ \bottomrule 
\end{tabular}}
\label{res:abfast}
\end{table}

\subsubsection{Study on Dynamic Frequency FFN}
We further verified the effectiveness of Dynamic Frequency FFN. As shown in Table \ref{res:abexp}, regardless of whether it is under the condition of Holistic Frequency Attention or Swin Attention, removing DF FFN leads to a decline in PSNR and SSIM, proving that the frequency-domain feed-forward network is effective. In addition, the visual comparisons in Figure 1 also demonstrate that the removal of DF FFN results in incorrect adjustment of image exposure conditions. This further confirms the importance of using a learnable matrix instead of a convolutional kernel in image restoration tasks.

\begin{table}[h]
\centering
\caption{Exploring the effect of Laplacian Pyramid Decomposition. R1 means replacing the U-Net Restorer with HDFformer in stage 1 of the pipeline.}
\resizebox{1\columnwidth}{!}{
\begin{tabular}{cccccccccc}
 \toprule
R-1         & R-2         & R-3         & R-4         & Lpls Decomposition      & PSNR$\uparrow$    & SSIM $\uparrow$   \\ \midrule
-         & -         & -         & -         & \ding{55} & 22.576 & 0.841  \\ \midrule
\ding{55} & \ding{55} & \ding{55} & \ding{55} & \ding{51} & 20.205 & 0.769  \\
\ding{51} & \ding{55} & \ding{55} & \ding{55} & \ding{51} & 21.682 & 0.813  \\
\ding{51} & \ding{51} & \ding{55} & \ding{55} & \ding{51} & 22.198 & 0.829  \\
\ding{51} & \ding{51} & \ding{51} & \ding{55} & \ding{51} & \textbf{22.812} & \textbf{0.859}  \\
\ding{51} & \ding{51} & \ding{51} & \ding{51} & \ding{51} & 22.809 & 0.857  \\ \bottomrule
\end{tabular}}
\label{ablpls}
\end{table}

\subsubsection{Study on Laplacian Pyramid Decomposition}
To further elucidate the efficacy of our proposed pipeline, we conducted an ablation study. Our pipeline is primarily divided into four stages. For baseline comparison, we initialized each stage with a standard U-Net architecture. The stages are then incrementally swapped out for HDFformer, our proposed module, in a hierarchical manner.

As shown in Table \ref{ablpls}, a progressive improvement in both PSNR and SSIM scores is observed as U-Nets were incrementally replaced by HDFformer modules. However, a saturation point is reached when all U-Nets were entirely substituted by HDFformer, showing a slight decline in the performance metrics.
This subtle degradation can be attributed to the overfitting tendencies of pure transformer architectures. Our findings corroborate that a balanced blend of convolutional layers and transformers, specifically the combination denotes as R1-R2-R3, leads to optimized performance.
Interestingly, when the Laplacian Pyramid Decomposition is removed, the performance remained commendable, though not as optimal as the full-fledged version. This highlights the robustness of HDFformer while also indicating that the Laplacian Pyramid Decomposition contributes to the superiority of the full model.

\section{Conclusions}
In this paper, we have proposed a spectral-domain attention and feed-forward network. We have validated their potential in both low-level and high-level tasks related to exposure, including exposure correction, low-light enhancement, multi-exposure fusion, low-light face detection, and low-light semantic segmentation. These tasks, which require sophisticated manipulation and understanding of image exposure, benefit significantly from our proposed spectral domain operations. Unlike traditional attention and feed-forward operations that operate in the spatial domain, our methods leverage the unique properties of the frequency domain to achieve superior performance with lower computational overhead.

\section{Acknowledgements}
This work is partially supported by China Postdoctoral Science Foundation (2023M730741), and the National Natural Science Foundation of China (Nos.62302078)

\bibliographystyle{unsrt}

\bibliography{FFT}

\end{document}